\documentclass[lettersize,journal]{IEEEtran}
\usepackage[dvipsnames]{xcolor}

\usepackage{todonotes}

\usepackage{amsmath,amsfonts,amssymb}
\usepackage{algorithmic,algorithm}
\usepackage[export]{adjustbox} 
\usepackage[noadjust]{cite}

\usepackage{lineno}
\usepackage{soul,color}
\usepackage{comment}

    
\makeatletter
\newcommand{\shorteq}{%
  \settowidth{\@tempdima}{-}
  \resizebox{\@tempdima}{\height}{=}%
}
\makeatother
\usepackage{multirow, hhline}
\usepackage{subfloat}
\usepackage{mathtools, mathdots}
\usepackage{hyperref}
\hypersetup{
    colorlinks=true,
    linkcolor=blue,
    filecolor=magenta,      
    urlcolor=blue,
}
\usepackage{caption, subcaption} 
\usepackage{textcomp}
\usepackage{gensymb}
\usepackage{diagbox}
\usepackage{enumitem}
\usepackage{siunitx}
\usepackage{lipsum}
\usepackage{optidef}
\usepackage{eso-pic}

\usepackage{caption}
\usepackage{makecell}

\DeclareMathAlphabet\mathbfcal{OMS}{cmsy}{b}{n}

\newcommand{\JP}[1]{{{\color{teal}JP: #1}}}

\begin{document}
\AddToShipoutPictureBG*{%
  \AtPageUpperLeft{%
    \setlength\unitlength{1in}%
    \hspace*{\dimexpr0.5\paperwidth\relax}
    \makebox(0,-0.75)[c]{\parbox{0.8\textwidth}{\centering \small This paper has been accepted for publication in the 7th Iberian Robotics Conference (ROBOT). Please cite as: Montano-Oliv\'an, L., Placed, J. A., Montano, L., and L\'azaro, M. T. (2024). From underground mines to offices: A versatile and robust framework for range-inertial SLAM. In 2024 7th Iberian Robotics Conference (ROBOT) pp. 1-8. IEEE. DOI: 10.1109/ROBOT61475.2024.10796903.}}%
}}

\title{From Underground Mines to Offices: A Versatile and Robust Framework for Range-Inertial SLAM
}



\author{

Lorenzo~Montano-Oliv\'an\textsuperscript{1},
        Julio~A.~Placed\textsuperscript{1},
        Luis~Montano\textsuperscript{2},
        and~Mar\'ia~T.~L\'azaro\textsuperscript{1}\\[0.5em]
        

{\normalsize
     \textsuperscript{1}Instituto Tecnológico de Aragón (ITA), Zaragoza, Spain. 
                    \{lmontano, jplaced, mtlazaro\}@ita.es \\
                   \textsuperscript{2}Instituto de Investigaci\'on en Ingenier\'ia de Arag\'on (I3A), Universidad de Zaragoza, Zaragoza, Spain.  montano@unizar.es
                    \vspace{-0.5cm}
}
       

\thanks{This work was partially supported by DGA\_FSE T73\_23R and EU Project HORIZON-CL4-2022-RESILIENCE-01 MASTERMINE (ID: 101091895).}

}
 

\maketitle
\begin{abstract}
    Simultaneous Localization and Mapping (SLAM) is an essential component of autonomous robotic applications and self-driving vehicles, enabling them to understand and operate in their environment. Many SLAM systems have been proposed in the last decade, but they are often complex to adapt to different settings or sensor setups. 
    In this work, we present LiDAR Graph-SLAM \mbox{(LG-SLAM)}, a versatile range-inertial SLAM framework that can be adapted to different types of sensors and environments, from underground mines to offices with minimal parameter tuning. 
    Our system integrates range, inertial and GNSS measurements into a graph-based optimization framework. 
    We also use a refined submap management approach and a robust loop closure method that effectively accounts for uncertainty in the identification and validation of putative loop closures, ensuring global consistency and robustness. 
    Enabled by a parallelized architecture and GPU integration, our system achieves pose estimation at LiDAR frame rate, along with online loop closing and graph optimization. 
    We validate our system in diverse environments using public datasets and real-world data, consistently achieving an average error below 20 cm and outperforming other state-of-the-art algorithms.
\end{abstract}


%
\IEEEpeerreviewmaketitle

\section{Introduction}

\IEEEPARstart{T}{he} availability of accurate environment models is essential for autonomous robots and self-driving cars to interact and navigate their environments.
In recent years, Simultaneous Localization and Mapping (SLAM) algorithms~\cite{cadena16} have evolved significantly in response to the demands of 3D reconstruction and online localization problems, becoming one of the most studied topics in the robotics community and leading to numerous implementations and applications across various domains.
Modern SLAM systems are based on probabilistic and graph theories; a rich graph representation that is used to express the variables of interest to be estimated (the robot locations over time) while accommodating the inherent uncertainty of the acquired observations in an error minimization problem. As these observations are associated with the graph, they can be concatenated to form a geometric representation of the environment, \textit{i.e.,}~a map. 
Both geometric and topological outcomes can be further exploited in subsequent tasks, such as navigation, autonomous operation, map-based localization, and multi-session applications. 

\begin{figure} [t!]
    \centering
    \includegraphics[width=.98\linewidth]{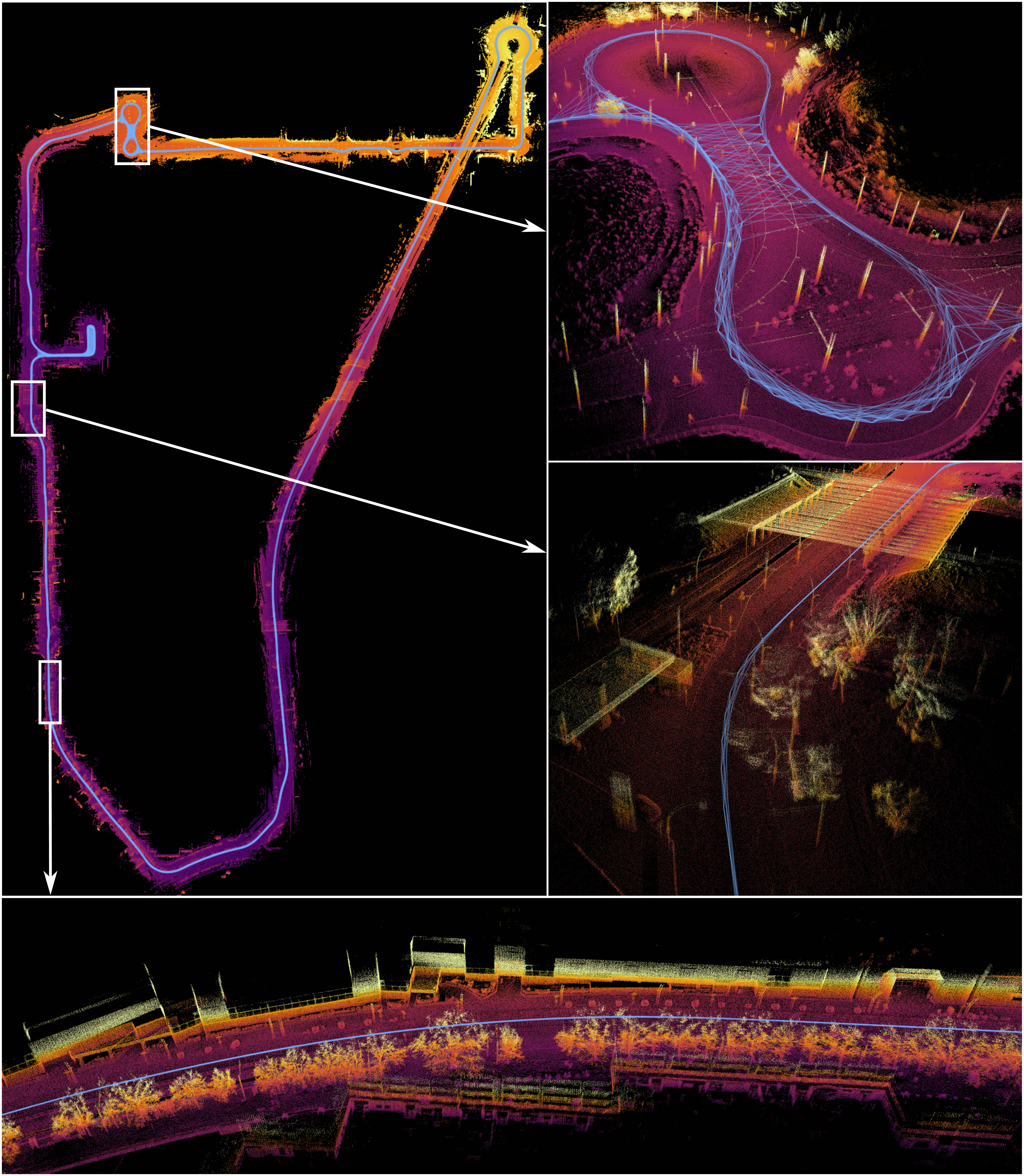}
    \caption{Visualization of the georeferenced 3D reconstruction and pose-graph (blue) built in an urban area ($\sim9$ km). LG-SLAM successfully operates in real time in a large environment, integrating heterogeneous measurements within a robust loop-closure validation process and producing accurate, consistent results even in GNSS-shadowed and featureless regions. The zoomed views in the figure highlight these key capabilities of our system.} 
    \label{fig:map_autotram}
\end{figure}

Autonomous robots and self-driving vehicle applications require algorithms to overcome challenges such as dust, rain, glare, low visibility, and other environmental phenomena. 
In such applications, methods that rely solely on visual sensors often lack robustness due to their sensitivity to lighting changes and glare. Despite their higher cost, Light Detection and Ranging (LiDAR) sensors have become indispensable in these applications, offering a significant advantage over visual-based methods in these scenarios due to their robustness and reliability against adverse conditions and their accuracy in measuring depth. Furthermore, there is a growing interest in the creation of accurate 3D reconstructions using handheld devices, particularly in areas such as construction sites where such technology can improve site visualization and progress monitoring by providing accurate models.

Typically, SLAM systems consist of odometry and mapping submodules. Many LiDAR Odometry (LO)~\cite{vizzo23} systems have emerged seeking to provide the robot with 
accurate pose estimation. However, unstructured and uniform environments, as well as aggressive motion, tend to compromise the performance and robustness of these single-sensor methods.
In response, the use of Inertial Motion Units (IMU) has become the standard~\cite{shan2020lio,chen23,xu22}. LiDAR-inertial Odometry (LIO) 
exploits the fact that IMU measurements are given at a much higher rate than LiDAR frames to: (i) correct for ego-motion distortion while acquiring a scan (intra-frame);
and (ii) constrain the motion between consecutive frames (inter-frame).
Although LIO systems can provide consistency over short- and mid-term operations, they are inevitably doomed to accumulate drift over time. 
Long-term consistency, crucial for a number of applications, requires loop-closing capabilities that associate spatially close but temporally distant visited regions. In addition, in outdoor applications, Global Navigation Satellite System (GNSS) measurements can be used to achieve global consistency and to georeference the 3D reconstructions. Together, loop closures and GNSS mitigate accumulated drift through a global optimization process.

Mapping processes face significant challenges and require adaptability to different environments and conditions. Existing methods often work effectively in structured environments, but struggle in more complex and unstructured environments such as underground or highly dynamic. 
Versatility to work with different sensor setups is another critical aspect. Several applications may require the use of different sensors, including different configurations of LiDAR (e.g., 16-channel versus 128-channel), RGB-D cameras, low-cost sensors, and setups with or without GPS. This versatility allows the mapping system to adapt to a wide range of scenarios and sensor configurations, increasing its robustness and applicability.

In this work, we present a robust and efficient LiDAR graph-SLAM system that is adaptable to a variety of environments and sensor configurations. 
Our system efficiently integrates some state-of-the-art techniques to achieve accurate and online performance, but it also introduces novel methods for submap handling (facilitating and robustifying data association between keyfames) and robust loop closing, as well as a consistent measurement integration using a two-stage graph optimization.  
The results of the experiments conducted in underground mines, urban scenes, and office-like environments with different sensor setups validate our approach and position our system at the forefront of the state-of-the-art.

\section{Related Work} \label{SII}

Point cloud registration forms the backbone of most LO and SLAM pipelines. It has been a relevant field for researchers since point-to-point Iterative Closest Point (ICP) was proposed in~\cite{besl1992method}, and many variants have emerged since then~\cite{rusinkiewicz2001efficient, pomerleau2015review}. 
However, the dense nature of point clouds still poses a significant computational challenge to modern algorithms. 
Consequently, many approaches attempt to reduce the size of the 3D cloud in order to optimize performance and efficiency.
On the one hand, \emph{feature-based methods} extract relevant characteristics of the scene. In~\cite{zhang2014loam, shan18, shan2020lio}, planar and edge features are extracted and then registered by computing point-to-plane and point-to-line distances within a voxel grid-based map representation. Methods based on the Normal Distribution Transform (NDT) algorithm~\cite{biber03}, such as~\cite{magnusson2015beyond, akai2017robust}, have demonstrated superior matching performance, but at the cost of strong dependence on the appearance of the environment and fine parameter tuning. 
On the other hand, \emph{dense methods}~\cite{xu22,reinke2022locus2} use all points of the scan for registration. However, in order to achieve real-time performance, they usually pre-process the scan with heavy downsampling, which can lead to information alteration and loss, and result in a loss of accuracy.
In~\cite{behley18,koide2021fastgicp}, this problem is addressed by using GPU acceleration to speed up the registration process without the need of significantly reducing the data size.

Odometry systems that rely solely on LiDAR can face challenges such as environmental sensitivity and suffer from distortion caused by aggressive rotations or rapid motion. 
Therefore, LiDAR is typically used in conjunction with other sensors, such as IMU and GNSS to address such issues. This is evidenced by the fact that modern LiDAR sensors typically integrate IMUs and provide the ability to synchronize its clock to the GNSS signal. 
The design of approaches using sensor fusion can typically be divided into two groups: loosely- and tightly-coupled. 
LOAM~\cite{zhang2014loam} and its variant LeGO-LOAM~\cite{shan18} are examples of the former group, where the IMU is used for motion correction and scan matching, but not in the optimization process.
On the contrary, 
LIO-SAM~\cite{shan2020lio} fall into the second category: LiDAR and IMU measurements are jointly optimized and a superior accuracy is achieved.
In addition, NV-LIO~\cite{chung24} extracts the normal vectors of the scans for robust scan matching in indoor environments, and proposes a viewpoint-based loop closure module to avoid wrong matches that are blocked by the walls. 
DLIO~\cite{chen23} proposes the construction of continuous-time trajectories for precise motion correction and a nonlinear geometric observer to ensure convergence in the state estimation. Xu \textit{et al.}~\cite{xu22} use an iterated Kalman filter instead to tightly couple the sensor measurements. 
In addition,~\cite{shan2020lio} and~\cite{koide2019portable} introduce absolute GNSS measurements in the optimization process to correct the accumulated drift.


Another crucial aspect of LIO systems is the use of submaps to generate a local map of the robot's surroundings. 
Submap generation provides robustness and compatibility with a wide range of sensors, particularly benefiting those with short field of view and LiDARs with sparse scan patterns.
The submaps mitigate the complexity of scan-to-map alignment operations and improve simpler and less robust scan-to-scan matching strategies.
In~\cite{shan2020lio}, the submaps contain features of a set of recent keyframes: planar and edge features from the current frame are aligned against the previous submap and integrated as the robot moves. 
Similarly, MULLS~\cite{pan21} keeps a local map containing static feature points from historical frames. 
In contrast, dense approaches often select submaps by cropping a region of the global map around the keyframes~\cite{reinke2022locus2}, resulting in higher computational complexity for loop closure search and less accurate submap alignment.

Recent odometry algorithms~\cite{chung24, chen23} achieve excellent performance by relying on a motion correction model and integrating additional sensors to make the system robust and efficient. 
However, these systems may accumulate drift and fail to achieve global consistency in long-term and large-scale problems. Loop closure detection is a critical component that addresses this problem by correcting for the accumulated drift when a region is revisited. Kim and Kim~\cite{kim18} propose ScanContext, a LiDAR-based place recognition mechanism to extract global descriptors. They can be used to estimate the transformation between the current scan and the revisited places. 
Recent learning-based methods such as PIN-SLAM~\cite{pan24} use neural point features in the local maps for loop closure detection, rather than using only the raw point cloud of a scan for feature extraction and matching. 
Ensuring the accuracy of the loop closures can help improve global consistency and prevent the system from diverging. For this reason, works such as~\cite{sunderhauf12, latif13} re-evaluate past loop closures to decide whether to update them with new information or to correct them if they contribute poorly to the graph consistency.

In this work, we present LG-SLAM, a versatile range-inertial SLAM system that can work in a wide range of scenarios, from underground environments to office settings. 
More specifically, our main contributions are as follows: 

\begin{itemize}
    \item \textit{Adaptive and modular design.} 
    Our system is highly adaptable and configurable to a wide range of sensors and robotic platforms. The parallelized architecture, coupled with GPU acceleration for 
    fast loop closing and tracking, ensures real-time operation.
    \item \textit{Submap management.}
    We effectively manage the local maps containing aggregated observations associated to specific locations, enhancing the overall accuracy.
    \item \textit{Robust loop-closing mechanism.} We employ a loop-closure checker module to evaluate the error in a window of closure candidates and selects those that minimize the overall error introduced in the optimization process.
    \item \textit{Global consistency.} GNSS integration in the mapping process ensures global consistency in outdoor settings and enables the creation of georeferenced maps.
    \item \textit{Superior performance.} Thorough experiments in public datasets and real-world scenarios demonstrate the performance of LG-SLAM, 
    achieving an accuracy of $10-20$~cm and outperforming other state-of-the-art algorithms.
\end{itemize}

\section{Methods}\label{SIII}

Let us consider a robot equipped with a LiDAR and, potentially, with an IMU and a GNSS antenna. Each sensor provides measurements at a certain rate and represents a complete coordinate system to which measurements are referenced, \textit{i.e.,}~$\{l\}, \{b\}, \{g\}$, respectively. For the sake of simplicity, the robot and GNSS frames will be assumed to be the same:~$\{r\}\equiv\{g\}$.
Each LiDAR scan $\tilde{\mathcal{L}}$ contains a set of 3D points acquired during the sweep time $\tau$. IMU provides acceleration ($\tilde{\boldsymbol{a}}$) and rotation rate ($\tilde{\boldsymbol{\omega}}$) measurements at high frequency. Finally, GNSS measurements contain global position information (\textit{i.e.,}~latitude, longitude and altitude). 

The $i$-th robot state will be given by the transformation between the robot and the map $\{m\}$ frames at that time:
\begin{equation}
    \boldsymbol{x}_i \equiv {}^m\boldsymbol{T}_{r_i}\triangleq \begin{pmatrix}{}^m\boldsymbol{R}_{r_i} &{}^m\boldsymbol{p}_{r_i} \\ \boldsymbol{0} &1\end{pmatrix}\in SE(3)\,,
\end{equation}
\noindent where ${}^m\boldsymbol{R}_{r_i}\in SO(3)$ and ${}^m\boldsymbol{p}_{r_i}\in\mathbb{R}^3$ are the rotational and translational terms, respectively. 
All the robot states up to the current time $t$ will be denoted as $\mathcal{X}\triangleq\{x_i\ |\ i=1,\dots,t \}$. Equivalently, the acquired observations will be $\mathcal{Z}\triangleq \{z_i \ | \ i=1,\dots,t\}$.
On the other hand, the map representation $\mathcal{M}$ at a certain moment will be encoded by a set of local maps, or \emph{submaps} $\mathcal{S}$, each of them associated to a robot state:
\begin{equation}
    \mathcal{M} \triangleq \{ \mathcal{S}_i \ | \ i = 1,\dots,t\} \,,
\end{equation}
\noindent with $\mathcal{S}_i$ containing a number of \emph{aggregated} LiDAR scans.

\begin{figure} [t!]
    \centering
    \includegraphics[width=.95\linewidth]{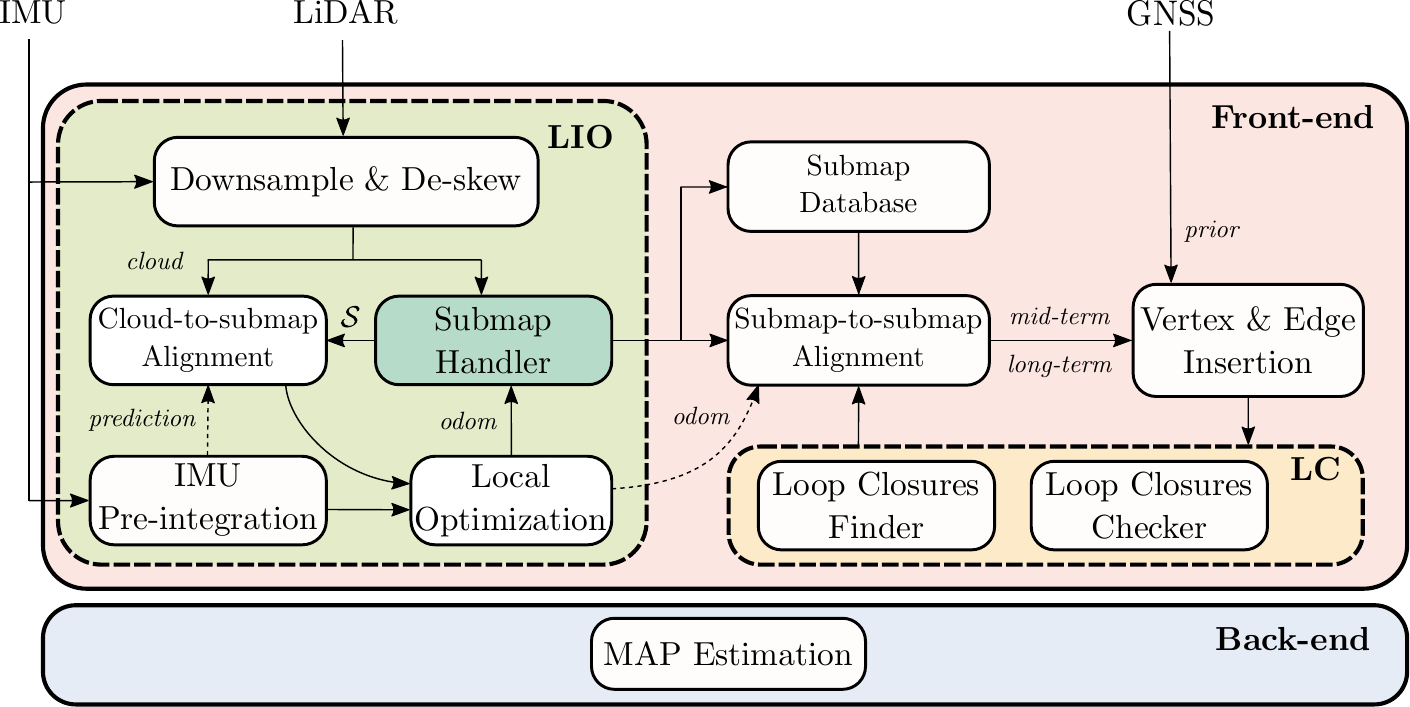}
    \caption{Overview of LG-SLAM.
    }
    \label{fig:overview}
\end{figure}

\subsection{System Overview}
The goal of our system is to estimate the optimal set of the current and past robot states,
\begin{equation}
    \mathcal{X}^\star \triangleq \underset{\boldsymbol{x}\in\mathcal{X}}{\mathrm{argmax}} \ p(\boldsymbol{x}\ |\ \mathcal{Z}) \,, \label{eq:opt_problem}
\end{equation}

and to build an accurate map representation of the environment; the latter can be obtained once the optimal robot states are known by correctly transforming and concatenating their associated LiDAR measurements~\cite{grisetti10}.

The architecture of the proposed method is shown in Fig.~\ref{fig:overview}. 
Heterogeneous measurements from IMU, GNSS and LiDAR are combined in a two-step graph-based optimization framework. 
First, range-inertial observations are fused to estimate relative measurements and build local representations of the environment. Second, this information and GNSS data are used to create a well-constrained pose-graph consistent both in the short and long terms. 
Our system runs in real time on ROS2 Humble~\cite{macenski2022robot} thanks to its parallelized architecture and the use of modern GPU-accelerated scan registration methods. It can handle different sensor configurations and LiDAR technologies (\textit{e.g.,}~GNSS denial, $16$ to $128$ resolution channels, mechanical and short-range solid-state) as well as a wide variety of environments with minimal parameter tuning. 
The handful of main parameters of our implementation are contained in Table~\ref{tab:params}.

\subsection{Feeding Suitable Input Data}
Efficient and accurate SLAM estimates can only be achieved if the raw sensor data is properly prepared. In this vein, we perform a preprocessing step on the recently observed sensor data to correct distortions, and ensure real-time and robust performance.
First, to avoid the precision degeneration of LiDAR observations over long distances and the occlusions near the sensor, the input scans $\tilde{\mathcal{L}}$ are clipped between $c_{\text{min}}$ and $c_{\text{max}}$, removing all points outside this range. 

\begingroup
\setlength{\tabcolsep}{4pt} 
\def\arraystretch{1.1}
\begin{table}[t!]
    \centering
    \scriptsize
    \begin{tabular}{l|l|l|l}
         \textbf{Module} & \textbf{Symbol} & \textbf{Value} & \textbf{Description}\\ \hline
         \multirow{3}{*}{Pre-processing} &$c_{\text{min}
         }$ &$0.5$ &Min. detection range (m) \\
         &$c_{\text{max}}$ &$100$ &Max. detection range (m) \\
         & $r_{\text{input}}$ & $\{0.03-0.20\}$ & Downsampling resolution (m)\\ \hline
         \multirow{3}{*}{Aligner} 
         &$-$ &\{FastGICP, NDTOmp\} &Registration method \\
         &$r_{\text{align}}$ &$\{1-5\}\cdot r_{\text{input}}$ &Registration voxel res. (m) \\
         &$d_{\text{max}}$ &$\{0.05-0.5\}$ &Max. corr. distance (m)\\ \hline
         \multirow{2}{*}{Odometry} &$r_{\text{submap}}$ &$\{0.03-0.12\}$ &Submap resolution (m) \\
         &$\ell$ &$\{1-10\}$ &Submap max. size (clouds) \\ \hline
         \multirow{3}{*}{SLAM} &$\Delta_{\text{trans}}$ &$\{1.0-3.0\}$ &Keyframe trans. thres. (m) \\
         &$\Delta_{\text{rot}}$ &$\pi/8$ &Keyframe rot. thres. (rad) \\
         &$v_{\text{local}}$ &$5$ &Local subgraph size    \end{tabular}
    \caption{Main parameters (and typical values) of LG-SLAM.}
    \label{tab:params}
\end{table}
\endgroup

\begin{figure} [t!]
    \centering
    \centering
    \includegraphics[width=.7\linewidth]{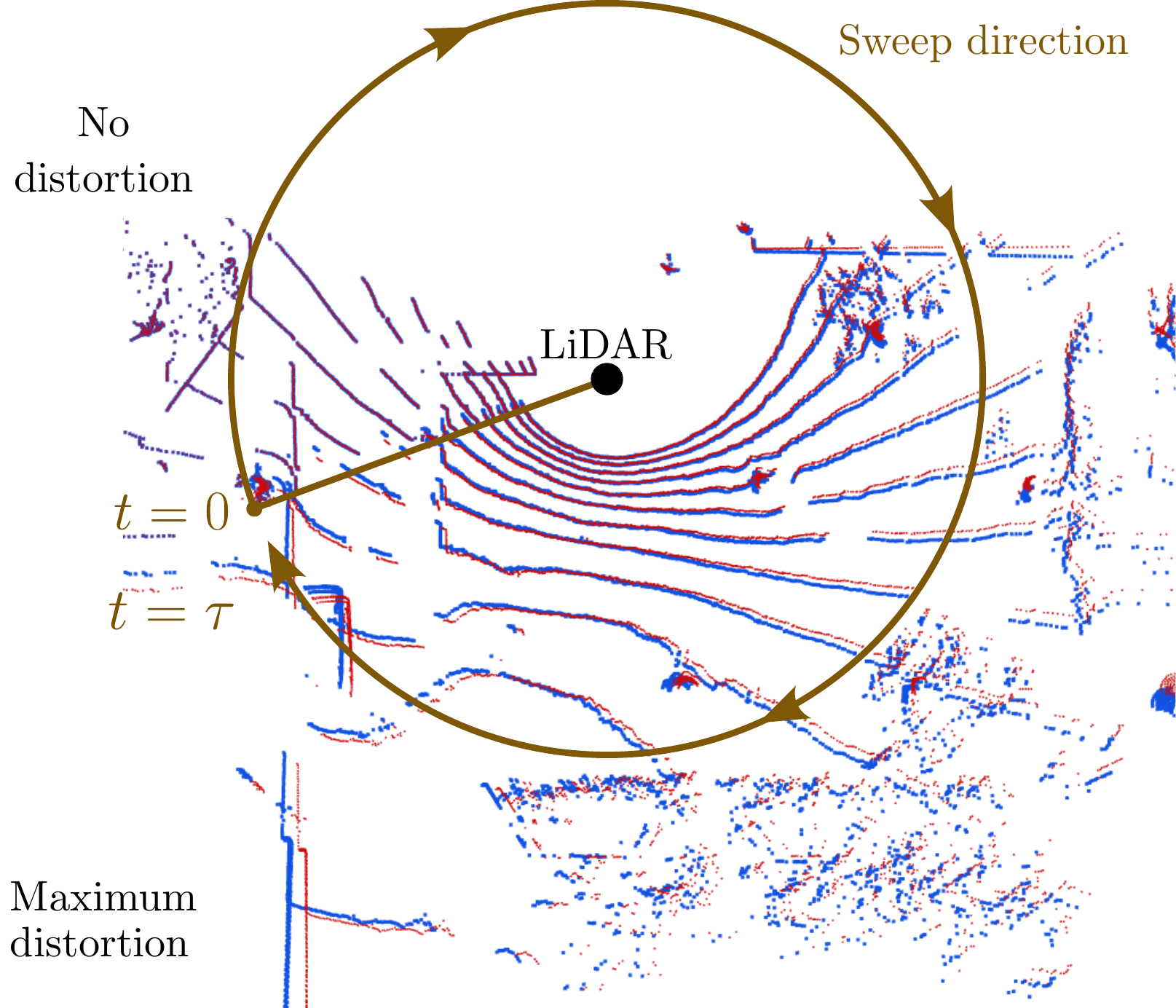} 
    \caption{Comparison of a scan before (blue) and after (red) de-skewing. Distortion increases with the sweep motion the and the point distance: it is maximum in the lower left region.}
    \label{fig:deskew}
\end{figure}

The distortion caused by the robot ego-motion during the acquisition of each scan can be reverted after estimating this motion. Each sweep consists of a set of range measurements continuously sensed during $\tau$. Despite $\tau$ tends to be small, the robot has usually moved away from the scan starting position and the distortion tends to be detrimental; this is specially relevant in handheld devices that allow for fast rotations and are prone to shaking.
We integrate the IMU angular velocity and acceleration to estimate the robot's motion over time and undistort the LiDAR scans. Given the high frequency of this sensor, this usually involves interpolation between IMU frames to obtain the motion between the start of the scan and the time a point is captured, although occasionally it may involve extrapolation. 
Figure~\ref{fig:deskew} shows the difference between an input scan before and after this process, known as \emph{de-sewking}.
Upon completion, $\tilde{\mathcal{L}}$ is voxel-downsampled to a desired resolution, $r_{\text{input}}$ (typically between $5$ and $10$ cm). Unlike other related work, our method does not require drastically downsampling the input cloud or sparsifying it as a set of features.

Finally, we transform the GNSS measurements to the Universal Transverse Mercator (UTM) coordinate system, obtaining the measured position of the antenna w.r.t. the origin of the UTM zone (\textit{i.e.,}~${}^{\text{UTM}}\tilde{\boldsymbol{p}}_{r_i}$ for time $i$). By storing the UTM position at the initial step ${}^{\text{UTM}}\tilde{\boldsymbol{p}}_{r_0}$, we can make all measurements relative to it and integrate them consistently with the rest of the observations, that is: ${}^{r_0}\boldsymbol{p}_{r_i}=({}^{\text{UTM}}\tilde{\boldsymbol{p}}_{r_0})^{\texttt{-}1} \ \cdot \ {}^{\text{UTM}}\tilde{\boldsymbol{p}}_{r_i}$. 
Subsequently, the initial UTM position can be used to recover a georeferenced position of the estimated robot states.

\subsection{Robust Odometry in a Sliding Graph Fashion}

Short-term consistency in SLAM can be achieved by abstracting pairs of temporally close measurements into ego-motion estimates. The accuracy of this process is key to subsequent tasks, and can be significantly improved by fusing data from different sensors and by efficiently associating the current observation with the $k$-last, on a one-to-many rather than a one-to-one basis. Figure~\ref{fig:overview} contains a block diagram of the proposed odometry module, which is described below.

The newly pre-processed LiDAR scan $\mathcal{L}_t$ can be aligned with respect to the previous submap $\mathcal{S}_{t\texttt{-}1}$ using a scan registration algorithm, such as ICP or NDT. Our system offers the flexibility to accommodate a variety of scan registration implementations, being the preferred ones the GPU-accelerated implementation of voxelized generalized ICP, FastGICP~\cite{koide2021fastgicp}, and the parallelized version of NDT, NDTOmp~\cite{ndtomp}.
The key parameters of the scan matching process are the voxel resolution, $r_{\text{align}}$, and the maximum distance between corresponding points, $d_{\text{max}}$. For NDTOmp, $r_{\text{align}}$ should be several times larger than $r_{\text{input}}$, while for FastGICP, it should be as close to it as the computational resources allow. The value of $d_{\text{max}}$ is environment dependent (it should scale with the distance between pairs of observed near points). 

\begin{figure}[t!]
    \centering
    \includegraphics[width=0.95\linewidth]{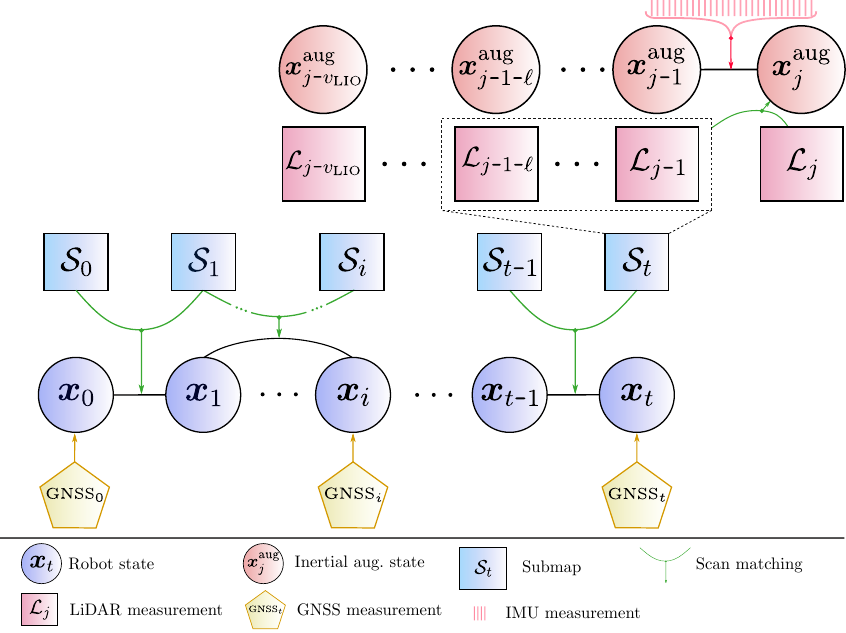}
    \caption{Graph architecture of LG-SLAM. Multiple sensor measurements are integrated into a pose-graph framework that combines pre-integration, GNSS, scan-to-submap and submap-to-submap matching constraints. Submap formation and loop closing are also represented.}
    \label{fig:graph_overview}
\end{figure}

The output of the above LiDAR Odometry (LO) process is combined with IMU measurements in a sliding graph optimization framework. Graph vertices are consistently introduced at LiDAR frame rate, encoding the augmented inertial state $\boldsymbol{x}^{\text{aug}}_t \triangleq (\boldsymbol{x}_t, \ \boldsymbol{v}_t, \ \boldsymbol{b}_t)$, where $\boldsymbol{v}_t\in\mathbb{R}^6$ is the linear and angular velocity, and $\boldsymbol{b}_t\in\mathbb{R}^6$ the gyroscope and accelerometer bias.
On the one hand, LO measurements are injected as vertex pose priors, with the covariance inherited from the registration method (see~\cite{koide2021fastgicp}) and taking into account the error between the matched point pairs. 
On the other hand, IMU measurements are preintegrated over $SE(3)$ to provide a condensed measurement between consecutive vertices. The motion model proposed in~\cite{forster16} in which the measurement noises are isolated is used to efficiently integrate raw measurements between LiDAR frames, therefore constraining their relative pose, velocity and bias variation.
To bound the optimization cost and to prune any degraded estimates, we reset the sliding graph every $100$ LiDAR frames: all the existing vertices are marginalized out, and a prior containing their information is connected to the first vertex of the new graph.
The graph optimization is handled by g2o~\cite{grisetti2011g2o} using the Levenberg–Marquardt algorithm and Huber robust kernels with $\delta=\chi^2(0.95)$ to smooth the process. The upper right portion of Figure~\ref{fig:graph_overview} (reddish colors) provides an overview of the LIO graph. 

High-frequency odometry (at IMU rate) is computed by composing the optimization result with the relative motion predictions from the preintegrated measurement model. This rough estimate can be used to initialize the next LO registration, boosting convergence.

The result of the optimization is also used to concatenate $\mathcal{L}_t$ to the previous submap forming $\mathcal{S}_t$. To keep tractability, each submap is limited to contain the last $\ell$ clouds, and downsampled to resolution $r_{\text{submap}}$ to prune duplicate points. 
Performing cloud-to-submap registration allows to maximize coverage compared to cloud-to-cloud matching; this improves the  performance and enables successful operation in confined environments and under aggressive motion, where the overlap between consecutive scans is limited.
In addition, the submaps allow to achieve high adaptability to different sensors (handling densities from $16$ to $128$ resolution channels) and different scan patterns by simply tuning $\ell$. 
For instance, for a high-resolution Ouster OS1-128 sensor (which can acquire tens of thousands of points per scan) $\ell$ will typically be between $1$ and $5$ to reduce the computational burden. However, for a Livox with a moving pattern, $\ell\geq8$ is used to stabilize the acquisition of points in the ground plane. 

\subsection{Keyframe Selection and Local Consistency}

To limit the computational complexity of the problem, robot state estimation and mid/long-term data association are not performed for every LiDAR frame, but only for those frames that contain useful information, so-called \emph{keyframes} (\textit{cf.} Eq.~\ref{eq:opt_problem}). 

We label a frame as a keyframe if it is separated from the previous one by more than $\Delta_{\text{trans}}$ meters or $\Delta_{\text{rot}}$ radians. 
Each keyframe will be a vertex in the pose-graph (framing the main optimization problem) and will encode the robot state $\boldsymbol{x}_t$ and a representation of the local environment given by the submap $\mathcal{S}_t$.
Since the inter-keyframe observations are accumulated in the submaps, we guarantee that no relevant information from the environment is lost.
In addition, each submap $\mathcal{S}_t$ is stored in a database, enabling efficient handling of large environments by dynamically loading and unloading specific submaps from memory and easing subsequent submap-to-submap matching. 

Instead of directly using the LIO estimates to constrain consecutive vertices, short/mid-term consistency is improved by aligning the submaps $\mathcal{S}_t$ and $\mathcal{S}_{t\texttt{-}1}$; this enforces $SE(3)$ pseudo-measurements between them with uncertainty derived from the matching process. See Figure~\ref{fig:overview} and the lower portion of Figure~\ref{fig:graph_overview}. The odometry prediction serves as initial guess for this registration.
In addition, GNSS data that is temporally close to a keyframe is introduced as a prior to further guide the optimization, with a covariance proportional to that given by the sensor (this value typically needs to be scaled up by several orders of magnitude to mitigate sensor overconfidence and be consistent with the rest of the observations). 

After adding a new keyframe and its associated short/mid-term constraints to the pose-graph, it is locally optimized as if it were a local bundle adjustment. 
The LIO estimate is used as the initial value for the last vertex, and a subgraph containing the newly added and the last $v_{\text{local}}$ nodes is optimized.
This optimization relies on the Levenverg-Marquardt algorithm, and pseudo-Huber robust kernels with $\delta=\chi^2(0.95)$.


\subsection{Global Consistency and Optimization}

Achieving globally consistent maps requires computing \textit{loop closures} to correct accumulated errors in the above processes by imposing constraints between geometrically close but temporally distant visited locations. These constraints are usually obtained through a registration process between the measurements gathered at each robot location. 
However, loop closure detection is a critical problem that must be handled carefully, as introducing an incorrect measurement can corrupt the optimization result and lead to an unrecoverable state. 
Additionally, it can be a computationally expensive task due to the potentially intensive search for candidate matches. 

Our system addresses long-term consistency in an efficient and robust manner. Loop closure search is executed in a parallel thread, allowing it to be easily decoupled from the main execution pipeline (see Figure~\ref{fig:overview}). Each time a new keyframe $\boldsymbol{x}_t$ is introduced into the graph, a search for potential loop closure keyframe candidates is performed based on the Mahalanobis and the graph distances. For each candidate $\boldsymbol{x}_j$, a submap-to-submap alignment is executed between $\mathcal{S}_t$ and $\mathcal{S}_j$. Given the good initial guess provided by the robust odometry and the use of submaps that ensures overlap, the convergence of the alignment is enhanced. 
As a result, we obtain a set of putative loop closure edges for $\boldsymbol{x}_t$ labeled with the estimated relative pose and a covariance matrix proportional to the alignment score. 
All candidate edges for the last $v_{\text{LC}}$ vertices are kept in a fixed-size queue to ensure robustness and coherence of the eventual choice (in practice, $v_{\text{LC}}=5$).
We then apply the voting scheme proposed in~\cite{lazaroIROS13} to discard outliers and to derive a subset of putative closures that minimizes the chi-squared error introduced into the graph. 
Finally, the selected loop-closing edges are added in bundles to the graph, which is then globally optimized. This ensures that there are no bottlenecks or permanent locking of the graph.

\section{Experiments}\label{SIV}
In order to demonstrate the accuracy and versatility of our approach, we provide validation results in two different setups. 
In the first case, our motivation is to position our method relative to other state-of-the-art algorithms. We have used three publicly available datasets that include urban and unstructured environments and that have been extensively used in related work. 
In the second setup, our motivation is twofold: to show the performance of the proposed method in real-world online settings, and to demonstrate its versatility and robustness. 
Overall, we have deployed our system in several real-world scenarios, covering a wide range of configurations: from indoor to large-scale outdoor experiments, from autonomous vehicles to handheld devices, from high-resolution sensors to more affordable alternatives.
All experiments presented in this section were performed on an Intel Core i7 CPU and an Nvidia GeForce RTX 3080 GPU.

\subsection{Dataset Benchmark}

\begin{figure}[t!]
    \centering
        \begin{subfigure}[t]{0.48\linewidth}
        \centering
        \includegraphics[max height=4.5cm,max width=\linewidth]{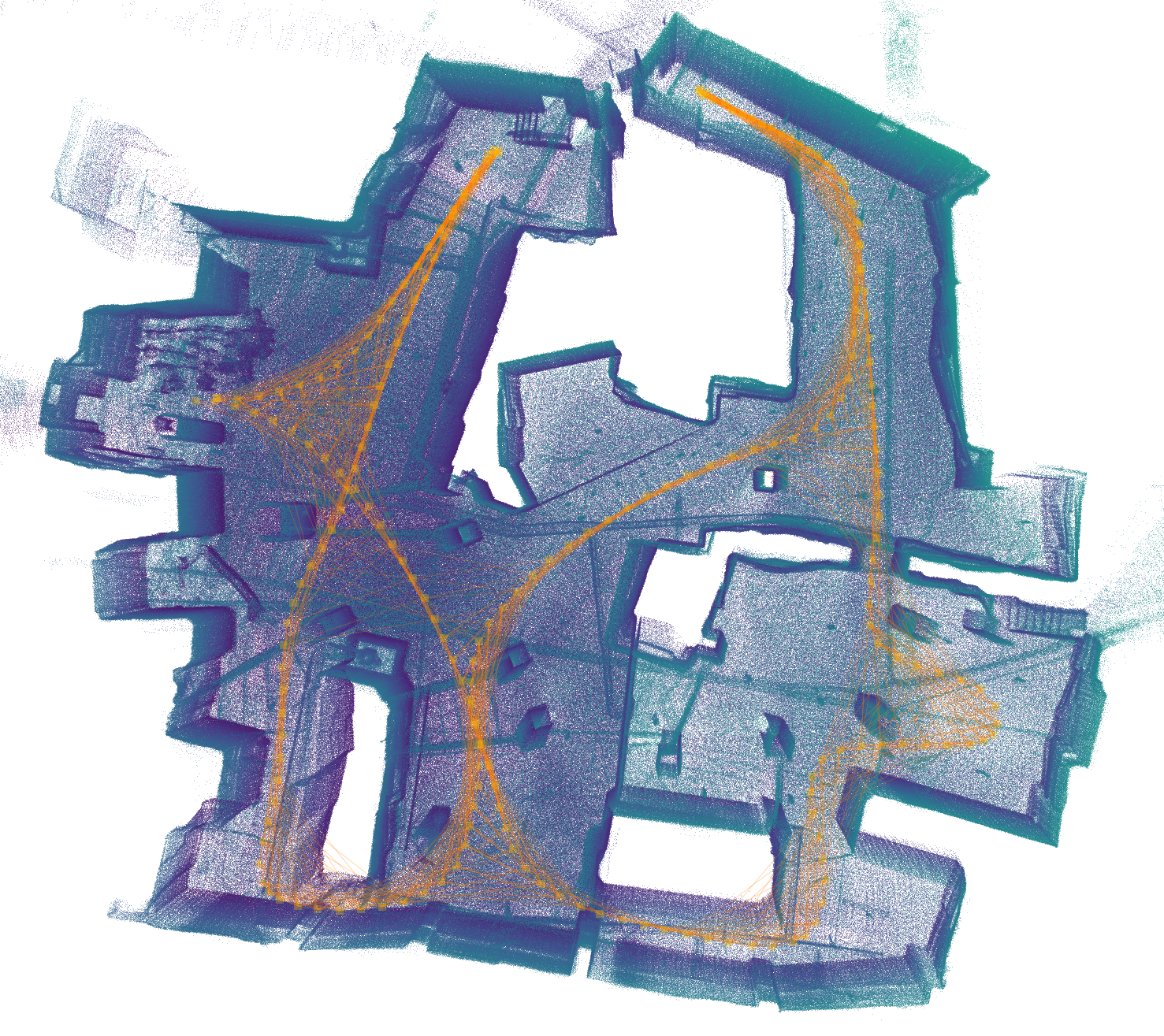} 
    \end{subfigure} \hfill
    \begin{subfigure}[t]{0.48\linewidth}
        \centering
        \includegraphics[max height=4.5cm,max width=\linewidth]{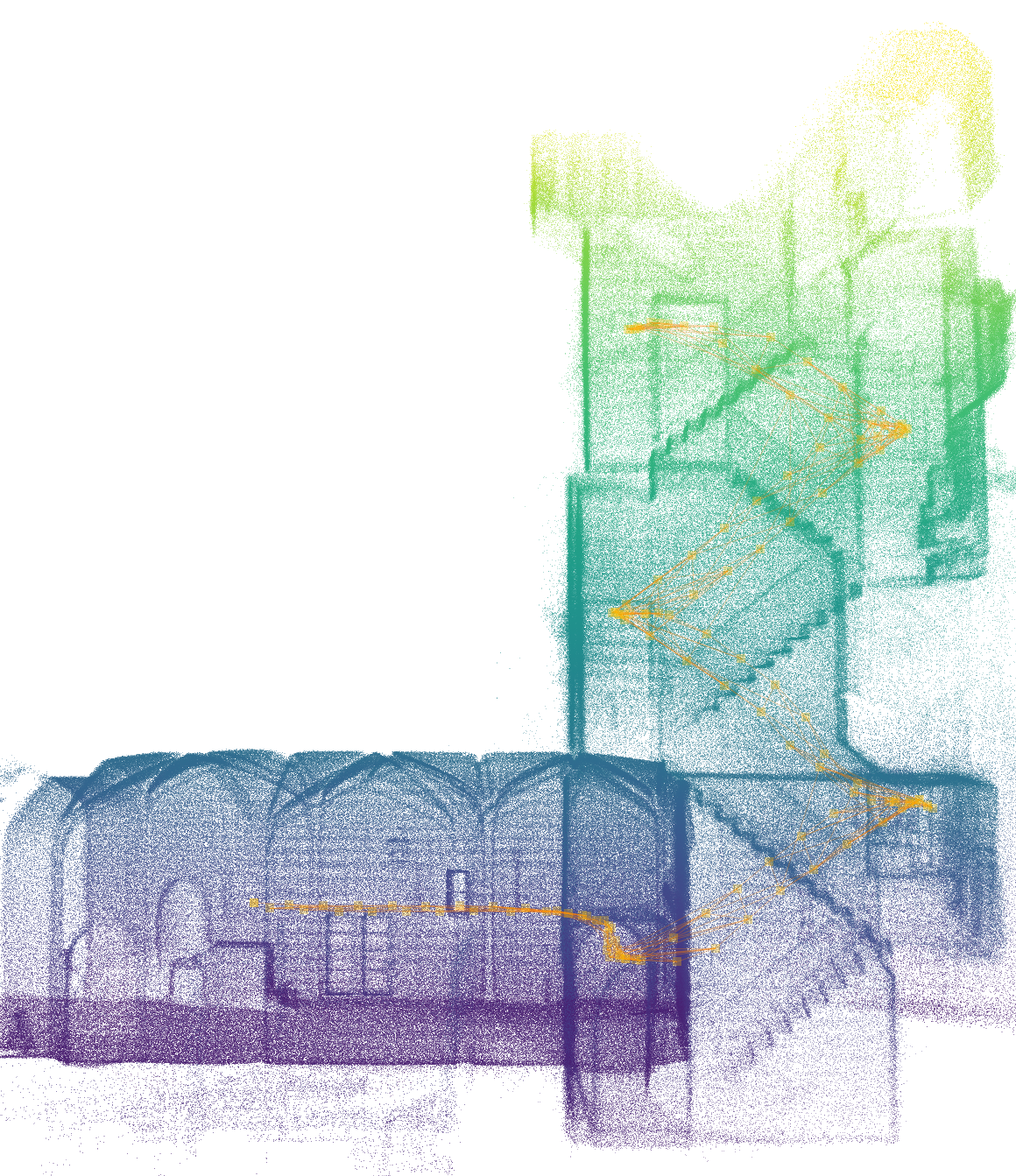} 
    \end{subfigure}
    \caption{Pose-graphs (orange) and 3D maps built by LG-SLAM in \emph{Underground} (left) and \emph{Stairs} (right) from NCE dataset~\cite{zhang21}.}
    \label{fig:map_nce}
\end{figure}

\emph{Newer College} (NC)~\cite{ramezani20} and \emph{Newer College Extension} (NCE)~\cite{zhang21} datasets contain different collections of data recorded using a handheld device in diverse environments. In the first case, data was recorded using an Ouster OS1-64 (10Hz) and its internal IMU (100Hz). In the second case, an Ouster OS0-128 (10Hz) and an Alpasense Core IMU (200Hz) were used. Ground-truth trajectories captured with a survey-grade 3D imaging laser scanner are provided.
We have evaluated our system using one sequence from the first dataset (\emph{short}) and five from the second one, namely three sequences from collections 1 and 3 (\emph{Quad}, \emph{Math}, \emph{Underground easy}), and two more complex sequences from collection 2: \emph{Cloister}, where visibility is limited, and \emph{Stairs}, where a three-story staircase is traversed. The length of the sequences ranges from $60$ m in \emph{Stairs} to $1.6$ km in \emph{Short}.
We have compared our system to three state-of-the-art LO and LIO algorithms (Fast-LIO2~\cite{xu22}, KissICP~\cite{vizzo23}, and DLIO~\cite{chen23}) and several SLAM systems (SuMa~\cite{behley18}, LeGo-LOAM~\cite{shan18} with improved loop closing using Scan Context~\cite{kim18}, LIO-SAM~\cite{shan2020lio}, F-LOAM~\cite{wang21}, MULLS~\cite{pan21}, MD-SLAM~\cite{di22}, PIN-SLAM~\cite{pan24} and NV-LIO~\cite{chung24})).
Table~\ref{tab:nce_benchmark} contains the Root Mean Squared error (RMSE) of the Absolute Trajectory Error (ATE), in meters, computed using \text{evo}~\cite{evo} with Umeyama alignment.
Results for the state-of-the-art algorithms were obtained from publicly available data in the recent works~\cite{pan24, chung24}.
Our method outperformed all other SLAM algorithms in the evaluated sequences. Only DLIO and NV-LIO achieved better results for two of the sequences, with an error difference below the centimeter level, and placing our algorithm as the second best. Remarkably, LG-SLAM successfully processed all sequences, while half of the evaluated methods struggled and even failed in the \textit{Stairs} sequence due to its intricate trajectory in a narrow indoor environment. This challenge was also reflected in algorithms that performed best on other sequences (such as Fast-LIO2 or DLIO, which lack loop-closing capabilities). In addition, we achieved the smallest average error across all sequences, which was $12$ cm.
Figure~\ref{fig:map_nce} shows the resulting 3D maps and pose-graphs (orange) for the \emph{Underground} (left) and the more complex \emph{Stairs} sequence (right), where the trajectory ATE is approximately $6$ and $5$ cm, respectively.

\begingroup
\setlength{\tabcolsep}{5pt} 
\begin{table}[t!]
    \centering
    \scriptsize
    \begin{tabular}{r|c|ccc|cc|c}
        \textbf{Method / Seq.} &Short &Quad &Math &Und &Cloister &Stairs &Avg.\\ \hline
        Fast-LIO2~\cite{xu22} &0.42 &0.083 &0.10 &0.11 &0.12 &0.11 &0.16 \\ 
        Kiss-ICP~\cite{vizzo23} &0.67 &0.096 &0.072 &0.33 &0.74 &$\times$ &0.38 \\ 
        DLIO~\cite{chen23} &\textbf{0.38} &\underline{0.073} &0.11 & 0.064 &0.094 &0.15 &0.15 \\ 
        \hline
        SuMa~\cite{behley18} &2.03 &0.28 &0.16 &0.09&0.20 &1.85 &0.77\\ 
        LeGO-LOAM~\cite{shan18} &- &0.09 &- &- &0.20 &3.20 &1.16\\ 
        LIO-SAM~\cite{shan2020lio} &\underline{0.40} &0.074 &0.081 & \textbf{0.058} & \textbf{0.074} &$\times$ &\underline{0.14}\\ 
        F-LOAM~\cite{wang21} &6.74 &0.40 &0.26 & 0.09&7.69 &$\times$ &3.04 \\ 
        MULLS~\cite{pan21} &2.51 &0.12 &0.35 &0.86 &0.41 &$\times$ &0.85\\ 
        MD-SLAM~\cite{di22} &- &0.25 &- &- &0.36 &0.34 &0.32 \\ 
        PIN-SLAM~\cite{pan24} &0.43 &0.09 &0.09 &0.07 &0.15 &\underline{0.06}&0.15 \\ 
        NV-LIO~\cite{chung24} &0.48 &0.076 &\textbf{0.062} &\underline{0.062} & 0.077 &0.087 &\underline{0.14} \\ 
        \hline 
        LG-SLAM (\emph{ours}) &\textbf{0.38} &\textbf{0.066} &\underline{0.070} & 0.064 &\underline{0.076} &\textbf{0.056} &\textbf{0.12}
    \end{tabular}
    \caption{ATE RMSE (m) in NC~\cite{ramezani20} and NCE~\cite{zhang21} datasets. Algorithms are sorted by most recent. Best results are bold and second best results are underlined. $\times$ indicates a failure, while $-$ indicates that results are not publicly available.}
    \label{tab:nce_benchmark}
\end{table}
\endgroup

\begin{figure} [t!]
    \centering
    \includegraphics[width=0.9\linewidth]{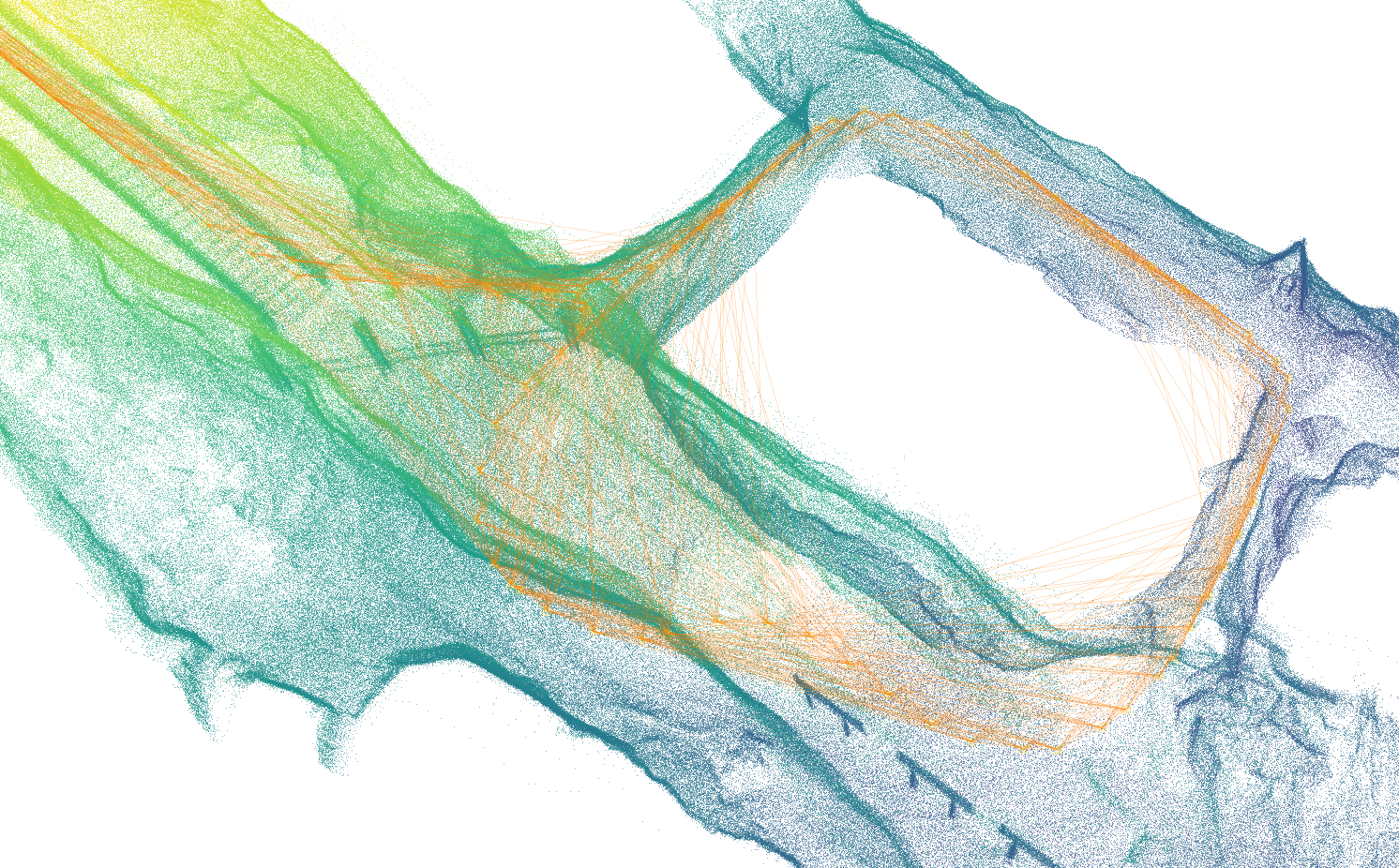}
    \caption{3D reconstruction and pose-graph (orange) generated by our algorithm in a section of \emph{Laurel Caverns} (SubT-MRS dataset~\cite{zhao24}).}
    \label{fig:map_subt}
\end{figure}

The \emph{SubT-MRS} dataset~\cite{zhao24} contains 3 years of data from the Defense Advanced Research Projects Agency (DARPA) Subterranean (SubT) Challenge and 2 more years of diverse environments. All sequences were recorded using a Velodyne VLP16 (10 Hz) and an Epson-G365 IMU (200 Hz), and ground-truth trajectories were derived from a high-precision scanner.
For comparison, we used six sequences (\emph{Urban}, \emph{Tunnel}, \emph{Cave}, \emph{Nuclear 1}, \emph{Nuclear 2} and \emph{Laurel Caverns}) that are part of the SLAM Challenge~\cite{zhao24} and for which detailed results of state-of-the-art methods are available. All sequences are recordings of real-world underground environments but contain different structures, degradation levels, and trajectories ranging from $120$ m to $1.4$ km. 
Table~\ref{tab:subt_benchmark} contains the ATE (RMSE) for each sequence. We report the average error of all methods participating in the challenge and the best result achieved by any team (notably, the best results were achieved by different algorithms depending on the sequence). Some teams used~\cite{xu22} or~\cite{chen23} as a backbone for LIO and incorporated loop closing capabilities using~\cite{kim18}, but we refer the reader to~\cite{zhao24} for details on the algorithms.
LG-SLAM outperformed every other method in $4$ out of the $6$ sequences, reducing the errors by more than $55\%$ compared to the best approaches. In the remaining two sequences, our approach remained well below the average and close to the best result.
We also report the overall performance in the last column of Table~\ref{tab:subt_benchmark}: the average error across all sequences for all teams was $83$ cm, with the best participant achieving an error of $27$ cm. Our method reduced the average error by more than four times to $19$ cm.
In Figure~\ref{fig:map_subt}, we present a zoomed-in region of the map created in \emph{Laurel Caverns} to provide a qualitative assessment of the 3D reconstruction.


Benchmarking the real-time performance and time consumption is a complex task because each method was run on a different platform (see~\cite{zhao24}).
For example, the best team in the SubT-MRS SLAM Challenge, who achieved $27$ cm ATE, required $51$ seconds per frame on an Intel i7 CPU (at $98\%$ load), but other candidates with reasonably bounded errors required only $6-125$ ms on similar platforms.
For comparison, LG-SLAM ran in real time on the platform described. The odometry module took $34\pm10$ ms per frame (less than the LiDAR rate) and mapping (including loop closing and graph optimization running in the background) took $190\pm100$ ms.

\begingroup
\setlength{\tabcolsep}{5pt} 
\begin{table}[t!]
    \centering
    \scriptsize
    \begin{tabular}{r|cccccc|c}
        \textbf{Method / Seq.} &Urban &Tunnel &Cave &Nuc1 &Nuc2 &Laurel & Avg. \\ \hline
        Average &0.63 &0.24 &0.69 &0.40 &0.61 &2.43 &0.83\\
        Best &\textbf{0.26} & \textbf{0.092} &\underline{0.62} &\underline{0.12} & \underline{0.22} &\underline{0.26} & \underline{0.27}\\ \hline
        LG-SLAM (\emph{ours}) &\underline{0.32} &\underline{0.13} &\textbf{0.37} &\textbf{0.055} &\textbf{0.14} &\textbf{0.15} & \textbf{0.19} \\
    \end{tabular}
    \caption{ATE RMSE (m) in SubT-MRS dataset~\cite{zhao24}. Best results are bold and second best results are underlined.}
    \label{tab:subt_benchmark}
\end{table}
\endgroup

\subsection{Real-world Deployment}

We have further evaluated the SLAM system in two different challenging real-world scenarios: a sensor-equipped car navigating in a large urban area and a person walking through an office building with a handheld recording device.

\begin{figure} [t!]
    \centering
    \includegraphics[width=.9\linewidth]{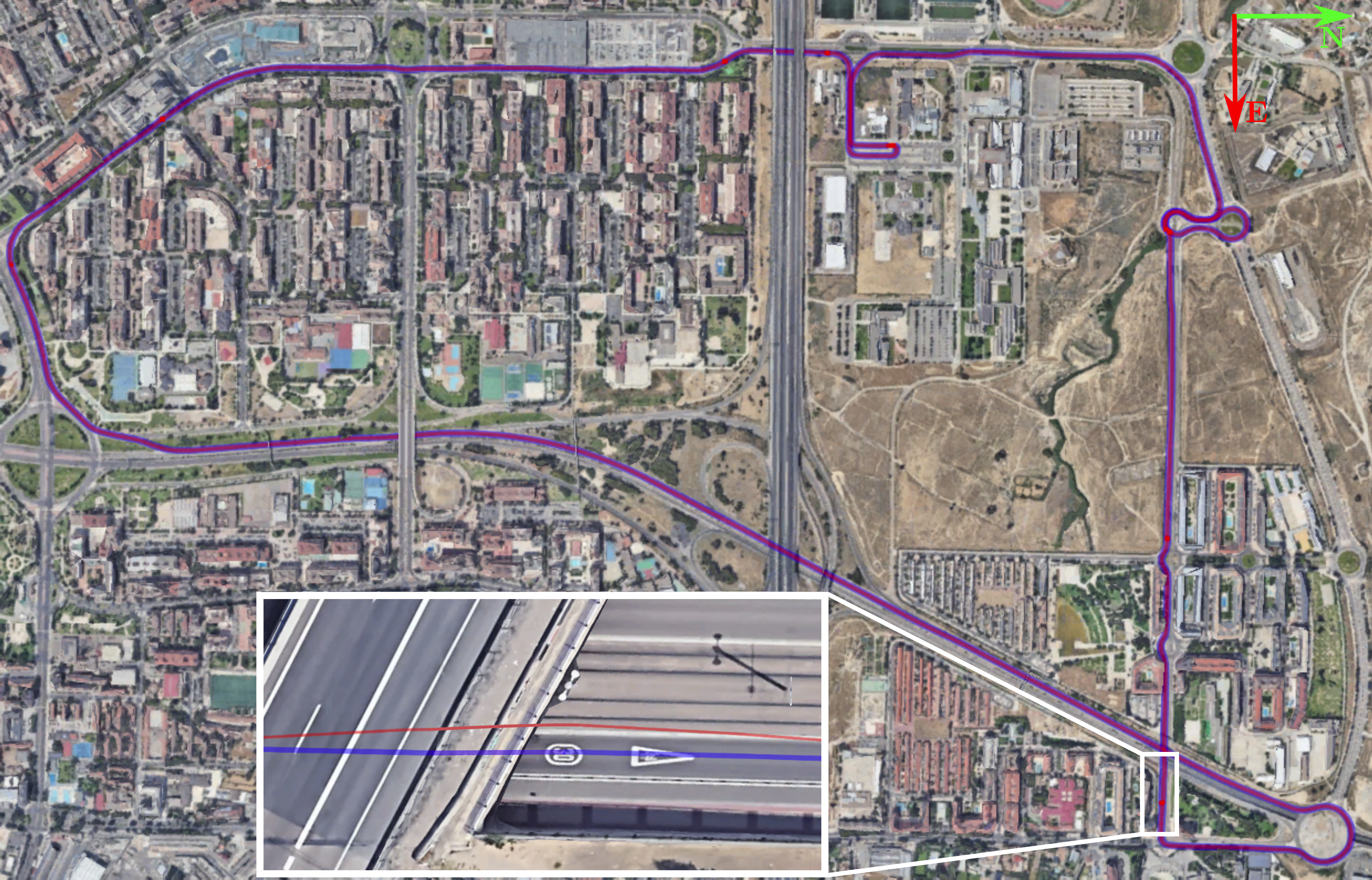}
    \caption{Georeferenced trajectory estimated by LG-SLAM (blue) and GNSS-RTK data (red), on top of a satellite view. The figure has been rotated $90$ deg clockwise for visualization purposes.}
    \label{fig:gps_autotram}
\end{figure}

In the first case, we used a car equipped with an Ouster OS1-128 sensor and an Ardusimple RTK2B GNSS antenna. The sequence covered a total length of $9$ km inside the city of Zaragoza (Spain), and included a variety of environments: open areas, large buildings, moving vehicles and pedestrians, etc.
The purpose of this experiment is to demonstrate the ability of our system to handle larger outdoor environments and to highlight its potential for georeferencing maps accurately. This underscores the system's reliability and its practical utility in autonomous driving scenarios.
Figure~\ref{fig:gps_autotram} contains the trajectory recorded by GNSS-RTK (red) and the estimated pose by our system (blue). Both paths overlap except in those regions in which GNSS signal was lost (see zoomed region).
Figure~\ref{fig:map_autotram} shows the pose-graph and map built, as well as three zoomed regions that show the precise and detailed reconstruction.

\begin{figure*} [t!]
    \centering
    \includegraphics[width=.92\linewidth]{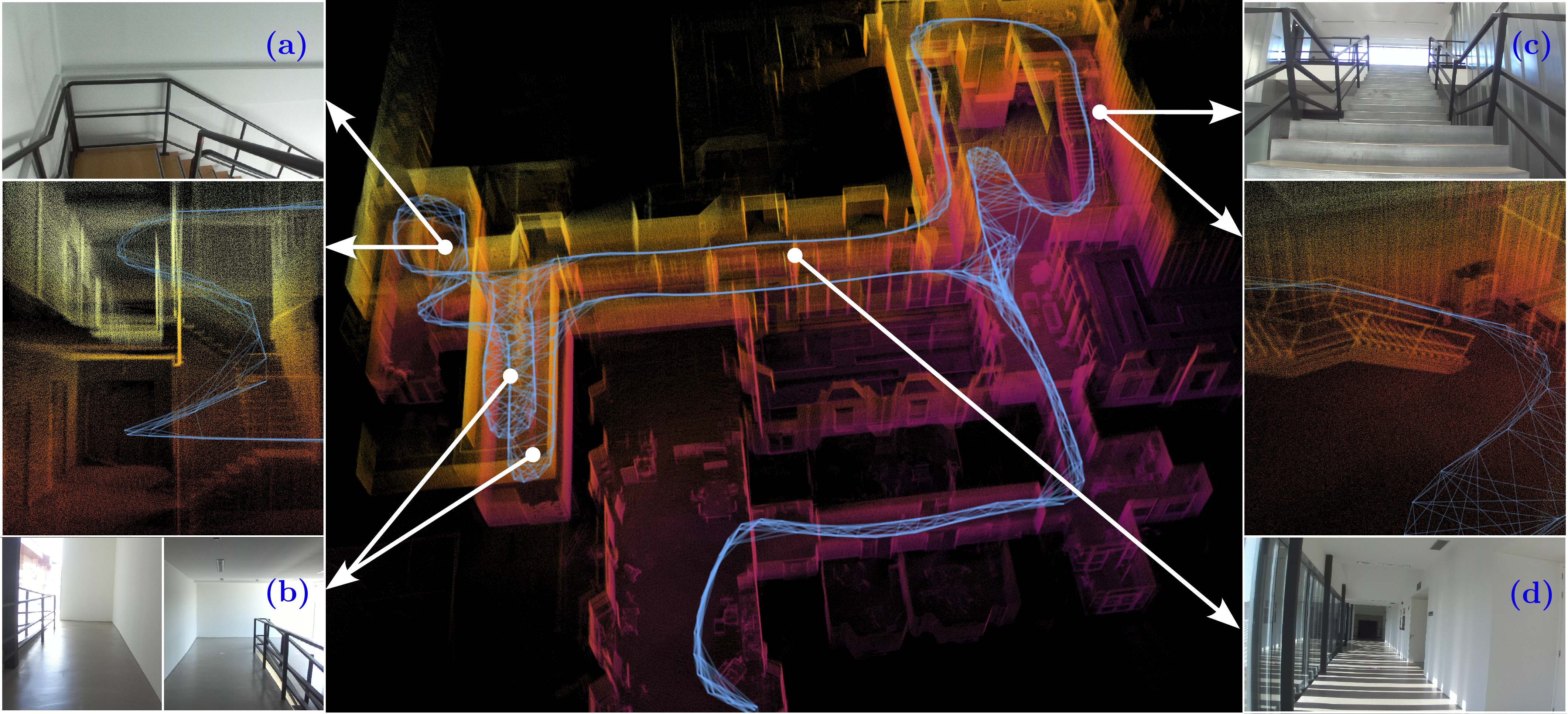}
    \caption{Complete 3D map and pose-graph (blue) generated in our facilities using a handheld device. In addition, specific viewpoints that represent a challenge to localization (\textit{e.g.,}~featureless or narrow regions, light reflections, etc.).}
    \label{fig:map_dora}
\end{figure*}

In the second case, we used our algorithm to reconstruct an indoor environment
using a handheld device equipped with an Ouster OS1-128.
The route covered a two-story, office-like building, and closed several loops over a span of $10$ minutes. This experiment showcases the versatility and robustness of our approach: the trajectory included rapid movements, shaking, challenging regions for the registration and loop closing algorithms, and areas with limited field-of-view. \mbox{LG-SLAM} successfully mapped the environment and accurately located the sensor on it.
Figure~\ref{fig:map_dora} shows the resulting accurate map and pose-graph (blue) of the entire building, including two zoomed-in regions. We also present recorded views to show some of the challenging regions in the environment and the scene diversity: (a) narrow stairwell where the sensor rotates rapidly ($\sim2$ rad/s), (b) potential false positive loop closing between floors, (c)
open hall with a staircase, and (d) geometric degradation and light reflections in a long corridor. 

\section{Conclusion}\label{SV}
In this work, we have introduced LG-SLAM, a versatile LiDAR-inertial SLAM system with a modular and adaptable design that enables mapping processes in diverse environments and accommodates a wide variety of sensor setups. 
The proposed map format goes beyond using solely metric information and leverages the graph structure. This format opens up research directions for long-term tasks such as map-based localization and map updating processes. 
The performance of our system is supported by thorough experiments in real-world scenarios.
We have benchmarked LG-SLAM against many popular SLAM algorithms on a variety of datasets, positioning our system at the state-of-the-art level.
Future work will also explore the use of additional range sensors (\textit{e.g.,}~radar) and the integration of new measurement sources (\textit{e.g.,}~visual odometry, prior knowledge from the environment). 

\ifCLASSOPTIONcaptionsoff
  \newpage
\fi

\bibliographystyle{IEEEtran}
\bibliography{bibliography}

\end{document}